\documentclass[sigconf]{acmart}
\usepackage{graphics}
\usepackage{adjustbox} 
\usepackage{float}
\usepackage{arydshln}
\usepackage{graphicx}
\usepackage{diagbox}
\usepackage{xspace}
\usepackage[toc,page]{appendix}
\usepackage{stfloats}
\usepackage{subfigure}
\usepackage{amsmath}

\usepackage{enumitem}
\setlist{leftmargin=3mm}
\usepackage[linesnumbered,ruled,vlined]{algorithm2e}

\newcommand{\minisection}[1]{\vspace{3pt}\noindent\textbf{#1.}}
\newcommand{\methodFont}{\textsf}

\newcommand{\metalight}{\methodFont{MetaLight}\xspace}
\newcommand{\dqn}{\methodFont{DQN}\xspace}
\newcommand{\intelli}{\methodFont{IntelliLight}\xspace}
\newcommand{\ddpg}{\methodFont{DDPG}\xspace}
\newcommand{\frap}{\methodFont{FRAP}\xspace}
\newcommand{\press}{\methodFont{PressLight}\xspace}
\newcommand{\colight}{\methodFont{CoLight}\xspace}
\newcommand{\maml}{\methodFont{MAML}\xspace}
\newcommand{\noblock}{\methodFont{FreeFlow}\xspace}
\newcommand{\our}{\methodFont{GeneraLight}\xspace}

\graphicspath{{figure/}}

\AtBeginDocument{%
  \providecommand\BibTeX{{%
    \normalfont B\kern-0.5em{\scshape i\kern-0.25em b}\kern-0.8em\TeX}}}

\copyrightyear{2020}
\acmYear{2020}
\setcopyright{acmcopyright}\acmConference[CIKM '20]{Proceedings of the 29th ACM International Conference on Information and Knowledge Management}{October 19--23, 2020}{Virtual Event, Ireland}
\acmBooktitle{Proceedings of the 29th ACM International Conference on Information and Knowledge Management (CIKM '20), October 19--23, 2020, Virtual Event, Ireland}
\acmPrice{15.00}
\acmDOI{10.1145/3340531.3411859}
\acmISBN{978-1-4503-6859-9/20/10}
\begin{document}
\fancyhead{}

% \pagestyle{plain}
% \let\oldfoot\footnotetextcopyrightpermission
% \renewcommand\footnotetextcopyrightpermission[1]{
%     \oldfoot{Submission to CIKM 2020}
% }

\title{GeneraLight: Improving Environment Generalization of Traffic Signal Control via Meta Reinforcement Learning}

\author{Chang Liu}
\email{only-changer@sjtu.edu.cn}
\authornotemark[1]
\affiliation{
  \institution{Shanghai Jiao Tong University}
  \city{Shanghai}
  \country{China}
  }
  
\author{Huichu Zhang}
\email{zhc@apex.sjtu.edu.cn}
\authornote{Both authors contributed equally to this research.}
\affiliation{
  \institution{Shanghai Jiao Tong University}
  \city{Shanghai}
  \country{China}
  }

  \author{Weinan Zhang}
\email{wnzhang@apex.sjtu.edu.cn}
\affiliation{
  \institution{Shanghai Jiao Tong University}
  \city{Shanghai}
  \country{China}
  }
 
   \author{Guanjie Zheng}
\email{gjz5038@ist.psu.edu}
\affiliation{
  \institution{The Pennsylvania State University}
  \city{University Park}
  \country{United States}
  }
  
    \author{Yong Yu}
\email{yyu@apex.sjtu.edu.cn}
\affiliation{
  \institution{Shanghai Jiao Tong University}
  \city{Shanghai}
  \country{China}
  }

\begin{abstract}
The heavy traffic congestion problem has always been a concern for modern cities. To alleviate traffic congestion, researchers use reinforcement learning (RL) to develop better traffic signal control (TSC) algorithms in recent years. However, most RL models are trained and tested in the same traffic flow environment, which results in a serious overfitting problem. Since the traffic flow environment in the real world keeps varying, these models can hardly be applied due to the lack of generalization ability. Besides, the limited number of accessible traffic flow data brings extra difficulty in testing the generalization ability of the models.
In this paper, we design a novel traffic flow generator based on Wasserstein generative adversarial network to generate sufficient diverse and quality traffic flows and use them to build proper training and testing environments. Then we propose a meta-RL TSC framework \our to improve the generalization ability of TSC models. \our boosts the generalization performance by combining the idea of flow clustering and model-agnostic meta-learning. We conduct extensive experiments on multiple real-world datasets to show the superior performance of \our on generalizing to different traffic flows.
\end{abstract}

\begin{CCSXML}
<ccs2012>
<concept>
<concept_id>10002951.10003227.10003351</concept_id>
<concept_desc>Information systems~Data mining</concept_desc>
<concept_significance>500</concept_significance>
</concept>
<concept>
<concept_id>10002951.10003260.10003277.10003281</concept_id>
<concept_desc>Information systems~Traffic analysis</concept_desc>
<concept_significance>500</concept_significance>
</concept>
</ccs2012>
\end{CCSXML}

\ccsdesc[500]{Information systems~Data mining}
\ccsdesc[500]{Information systems~Traffic analysis}

\keywords{Traffic Signal Control,
Environment Generalized Model,
Meta Reinforcement Learning, Generative Adversarial Network}
\maketitle

\section{introduction}
With the rapid development of cities, the traffic congestion problem has become more and more severe, which significantly affects people's daily lives. Optimizing traffic signal control (TSC) systems is an important way to alleviate traffic congestion \cite{mirchandani2001real}. 

\begin{figure}[!bp]
    \centering
    \includegraphics[width=0.35\textwidth]{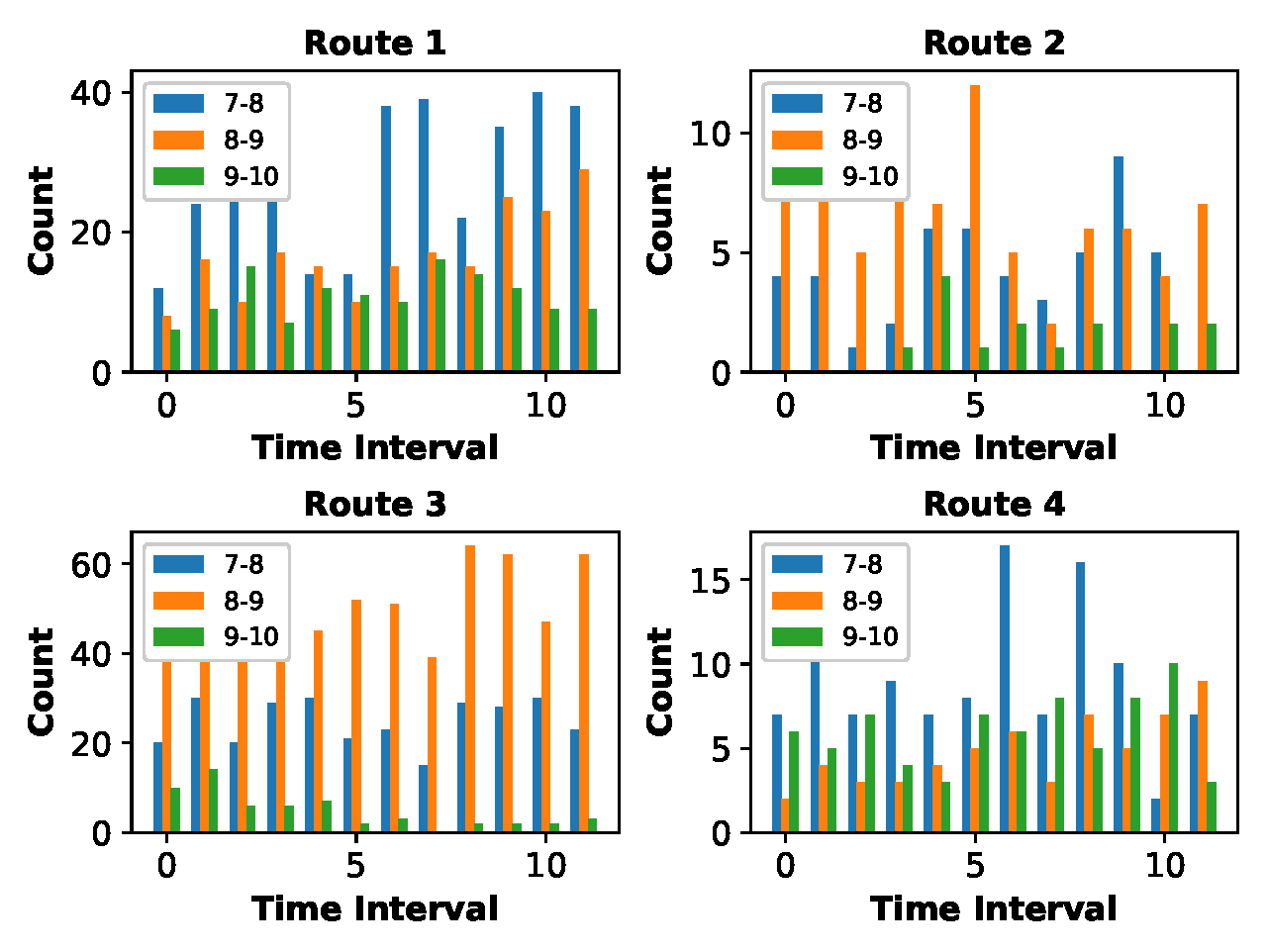}
    \caption{Traffic flow during three time periods (7-8a.m.;8-9a.m.;9-10a.m.) for a single intersection in  Hangzhou, China. For each route, we count the number of vehicles in each time interval (10 minutes).}
    \label{fig:real_flow_3}
    % \vspace{-20pt}
\end{figure}

Regarding TSC as a sequential decision-making problem, researchers have recently started to explore the TSC solutions with reinforcement learning (RL), and have achieved promising results \cite{wei2018intellilight, zheng2019learning, wei2019colight, xiong2019learning, wei2019presslight, chachatoward}.
Like the solutions of RL for game playing \cite{Mnih2015HumanlevelCT}, these methods train the TSC policy in a traffic simulator with a traffic flow, and test the solution using the same traffic flow just like playing the same game. However, the real-world situation is much more complicated as the traffic flow is always changing. Figure \ref{fig:real_flow_3} shows the real traffic flow of a single intersection in Hangzhou, China during three time periods. For each route, we count the number of vehicles in each time interval. As can be easily observed, the traffic flow varies a lot for different time periods. Obviously, under current settings, the aforementioned models are prone to overfitting to one traffic flow environment, and their performance may severely deteriorate if the traffic flow environment changes. This lack of generalization ability brings great difficulties of deploying these models to the real-world.

The only existing RL-based TSC algorithm that considers the generalization ability is \metalight \cite{Zang2020MetaLightVM}. \metalight aims to find a generalized model for any type of intersection and phase settings. For example, \metalight can be trained on a four-way intersection environment while being tested on a three-way intersection. However, instead of different state and action spaces which can be solved with more training time, different traffic flow environments is more critical and practical. The deployed TSC model must have the ability to handle different patterns of traffic flow (heavy or light, uniform or non-uniform).

In this paper, we aim at increasing the generalization ability of RL-based TSC algorithms to different traffic flow environments. To achieve this goal, we need to overcome two main challenges. First, the validation of a TSC algorithm may need a large amount of traffic data, whereas we can only acquire limited traffic data collected in the real-world due to the privacy and device accuracy issues. For example, in the single intersection data collected in Hangzhou, China\footnote{https://traffic-signal-control.github.io/\#open-datasets}~\cite{Wei2019ASO, zheng2019learning, wei2019colight}, we only have eleven one-hour real traffic flows. If we simply test our model on all these eleven traffic flow environments, our model may just overfit to these environments. Another challenge is how our model can quickly adapt to different traffic flows as the flow distribution in real-world may vary a lot from regular time to rush hours, from one region to another. We hope the TSC model can quickly adapt to different traffic flow environments, especially those that the model has not been trained on.

In this paper, to address the limited data issue, 
we design a novel traffic flow generator based on Wasserstein generative adversarial network (WGAN) \cite{Arjovsky2017WassersteinG} to generate sufficient diverse and quality traffic flows within various similarity levels as the real-world traffic. Based on the generated traffic flows, we can set up traffic environments for training and testing. Furthermore, to improve the generalization ability, we propose \our, an enhanced model-agnostic meta-learning (MAML) framework. \our combines the idea of flow clustering~\cite{MacQueen1967SomeMF} and the MAML-based meta reinforcement learning algorithm~\cite{Finn2017ModelAgnosticMF} which learns prior knowledge from the traffic flow environments in the training set to better adapt to the new traffic flow environments in the test set.

In summary, the technical contributions can be listed as follows.
 \begin{itemize}
     \item To the best of our knowledge, we are the first to improve the generalization ability of TSC models which aims to make TSC models generalize to different traffic flow environments.
     \item We design a WGAN-based flow generator to resolve the limited real data issue and  build the train set and test set as a proper way to test the generalization ability of TSC methods.
     \item We propose \our to improve the generalization ability of TSC models, which combines the idea of flow clustering and meta RL. \our can quickly adapt to different flow environment.
 \end{itemize}
% We conduct extensive experiments on multiple real-world datasets to verify the generalization ability and effectiveness of \our over strong baselines. 
\section{Related Work}
\minisection{Reinforcement Learning for Traffic Signal Control}
Traffic signal control is an important topic in transportation field for decades, which aims to optimize people's travel time by rescheduling the traffic signal plans. People in transportation fields have proposed many methods that highly rely on prior human knowledge~\cite{roess2004traffic,  mirchandani2001real, miller1963settings}. Recently, using reinforcement learning to control the traffic signal has become a popular topic~\cite{mannion2016experimental, wiering2000multi}. Learning-based traffic signal control method is more general in application since it does not require any manually predefined signal plans. Earlier learning-based works mainly utilize the idea of tabular Q-learning with discrete state representations~\cite{abdulhai2003reinforcement, el2010agent}.

Since the idea of deep reinforcement learning was proposed, people start to apply deep reinforcement learning methods to the TSC problem. IntelliLight~\cite{wei2018intellilight} first uses DQN~\cite{Mnih2015HumanlevelCT} to solve TSC problem. FRAP~\cite{zheng2019learning} focuses on improving model performance by reducing the exploration space. CoLight~\cite{wei2019colight} proposes to use the multi-head attention~\cite{velivckovic2017graph} to coordinate the traffic signals in multi-intersection environment. DemoLight~\cite{xiong2019learning} proposes to let reinforcement learning agents learn from some simple models to speed up the training process. PressLight~\cite{wei2019presslight} redesigns the representation of state and reward, and get good performance in both single and multiple intersection environments. MPLight~\cite{chachatoward} successfully conduct experiments on the super-large road network of 2510 intersections.

However, the aforementioned RL methods have one common problem that they did not consider the generalization in different traffic environments. In most papers, they train their agents for many rounds in one environment and test their agents in the same environment. MetaLight~\cite{Zang2020MetaLightVM} is the only exception which aims to train and test one agent for different intersections, such as train one agent on a four-way intersection and test it on a five-way intersection. MetaLight mainly focuses on the difference of action and observation spaces. In this paper, we aim to design a TSC model that can be generalized to different traffic flow environments.

\minisection{Meta Reinforcement Learning} Meta reinforcement learning aims to solve a new reinforcement learning~\cite{Sutton1998ReinforcementLA} environment by utilizing the experience learned from a set of environments. Early approaches use a recurrent network to learn from the experience in the training environments, such as RL2~\cite{Duan2016RL2FR} and \cite{Wang2017LearningTR}. Then, SNAIL~\cite{Mishra2018ASN} combines the idea of wavenet~\cite{Oord2016WaveNetAG} and RL2 to better utilize historical information. EPG~\cite{Houthooft2018EvolvedPG} uses a meta loss function to store the information of different environments. MetaGenRL~\cite{Kirsch2019ImprovingGI} improves EPG by redesign the meta loss function by gradient descent to save training time. MAESN~\cite{Gupta2018MetaReinforcementLO} adds noise as a part of the state to enhance the exploration ability of policy. Some people think robust optimization~\cite{Kuhn2019WassersteinDR, Jafra2019RobustRL} is also a possible solution. EPOPT~\cite{Rajeswaran2017EPOptLR} searches the environment in train set to find some hard environments, and optimize the policy on the hard environments. RARL~\cite{Pinto2017RobustAR} uses adversarial networks~\cite{Goodfellow2014GenerativeAN} to improve the robustness of model. People have recently proposed novel unsupervised learning-based models to improve the generalization ability~\cite{Eysenbach2018DiversityIA, Gupta2018UnsupervisedMF}. 
% As far as we know, there is currently no meta reinforcement learning method for multi-agent reinforcement learning.

MAML~\cite{Finn2017ModelAgnosticMF} is the most popular meta-learning algorithm that aims to optimize the global parameter initialization, which significantly
improves the effectiveness and efficiency of reinforcement learning in new environments. However, people find that only one global initialization is not enough for diverse environments: MUMOMAML~\cite{Vuorio2018TowardMM} uses environment-specific global initialization for different environments. HSML~\cite{Yao2019HierarchicallySM} aims to tailor the global initialization to each environment, by a hierarchical structure that combines the embedding of environments and global initialization. In this paper, we follow the tailoring idea and propose \our, which combines the idea of flow clustering to better tailor the global initialization to different environments.
\section{Preliminaries}
We first explain some key concepts in traffic signal control problem, which are widely accepted by previous work~\cite{wei2018intellilight,Wei2019ASO,wei2019presslight,zheng2019learning,wei2019colight,chachatoward}.
\begin{itemize}
    \item \textbf{Roadnet.} A roadnet $G$ is abstracted from some parts of the real city, including $N$ intersections and a number of roads connecting the intersections. Each road consists of several different lanes.
    \item \textbf{Traffic Movement.} For each intersection $i$, we denote the set of incoming lanes and outgoing lanes of an intersection as $L_{in}$ and $L_{out}$ respectively. A traffic movement is defined as the traffic traveling across an intersection from one incoming lane $l_{1}$ to an outgoing lane $l_{2}$, which we denote as ($l_{1}$,$l_{2}$).
    \item \textbf{Phase.} Traffic signals control the corresponding traffic movements, with the green signal indicating the movements are allowed and the red signal indicating the movements are prohibited. We denote the signal of movements as $a(l_{1}, l_{2})$, where $a(l_{1},l_{2}) = 1$ indicates the green light is on for movement $(l_{1}, l_{2})$, and $a(l_{1},l_{2}) = 0$ indicates the red light is on for movement $(l_{1}, l_{2})$. Finally, a phase \cite{zheng2019learning} is defined as the combination of several signals, as $\phi = \left\{(l_{1}, l_{2})|a(l_{1},l_{2}) = 1\right\}$, where $l_{1} \in L_{in}$ and $l_{2} \in L_{out}$. 
    \item \textbf{Traffic Flow.} Traffic flow $\mathbf V$ consists of vehicle movements. For each vehicle $i$, we have $r_i$ which is the route it follows and departure time $t_i$. To get a global picture of each traffic flow, we define $R$ as the set of different routes and partition departure time $t$ into $T$ time intervals and use a two-dimensional $|R| \times T$ vector to define the statistics of a particular traffic flow $\mathbf V$ where we count the vehicles of each route $r$ in each time interval $t$, shown as
    \begin{equation}
    \begin{split}
    \mathbf V &= [\mathbf V^{1}, \mathbf V^{2}, \cdots, \mathbf V^{|R|}]~, \\
    \mathbf V^{r} &= [V^{r}_{1},V^{r}_{2}, \cdots, V^{r}_{T}]~.    
    \end{split}
    \label{eq:flow}
    \end{equation}
    In most previous RL-based TSC methods, traffic flow is unchanged and remains the same when training and testing.
    
    % \begin{equation}
    % \begin{split}
    % F  = [F^{1}, F^{2}, \cdots, F^{|R|}]= [ &[F_{1}^{1}, F_{2}^{1}, \cdots, F_{T}^{1}], \\
    %       &[F_{1}^{2}, F_{2}^{2}, \cdots, F_{T}^{2}], \\
    %       &\cdots , \\
    %       &[F_{1}^{|R|}, F_{2}^{|R|}, \cdots, F_{T}^{|R|}] ]\\
    % \end{split}
    % \label{eq:flow}    
    % \end{equation} 
    \item \textbf{Traffic Environment.} Based on given traffic flow, the traffic environment will push vehicles into the roadnet at the corresponding departure time, and simulate vehicles travelling in the roadnet following its route. TSC algorithms can control each traffic signal according to current traffics. 
    \item \textbf{Distribution of Traffic Flows.} The traffic flow is always changing in the real world. Therefore, we regard the set of traffic flows as samples of a traffic flow distribution $D$. In particular, we regard the traffic flows in our dataset as the samples from the traffic flow distribution $D_{0}$ ($0$ denoting zero Wasserstein distance which will be explained soon).
    \item \textbf{Distance between Traffic Flow Distributions.} We use Wasserstein distance~\cite{Arjovsky2017WassersteinG} to define the difference between a new traffic flow distribution $D_{d}$ and $D_{0}$, as Equation~(\ref{eq:w_dis}) shows
    \begin{equation}
    \begin{split}
        \text{W-dis}(D_{d}) = W(D_{d}, D_{0}) = \inf\limits_{\xi \in \prod(D_{d}, D_{0})} \mathbb{E}_{(x, y) \sim \xi}[\|x - y\|]~.
    \end{split}
    \label{eq:w_dis}    
    \end{equation} 
    where $\prod(D_{d}, D_{0})$ denotes the set of all joint distributions $\xi(x,y)$ whose marginals are respectively $D_{d}$ and $D_{0}$.
    % \weinan{1. I think it is necessary to explain $\gamma$ and $\prod(D_{k}, D_{d})$. 2. $\gamma$ is heavily abused in Eq 2 5 and 9.}
\end{itemize}

\noindent\textbf{Problem Setting.} In this paper, we focus on improving the generalization ability of TSC methods to different traffic flows. We first generate traffic flow distributions that have different Wasserstein distance between $D_{0}$ with the help of WGAN. We use subscripts to show different W-distance between $D_0$ of a new traffic flow distribution (e.g. $D_{0.1}$). We will introduce the detail of this WGAN module in Section 4.1. Agents are trained in the traffic environments using traffic flows sampled from $D_{0}$, and are tested in new traffic environments using traffic flows generated from $D_{test}$. The goal of agents is to minimize the average travel time (explained in \ref{sec:metric}) of all vehicles on both training and testing traffic flow distribution.
\section{Method}\label{sec:method}
In this section, we will introduce the main modules of our proposed method. First, we propose a WGAN-based model to generate the train and test traffic flow environments for our RL model. WGAN is known to be one of the state-of-the-art methods for fitting data distribution and generating new data \cite{Arjovsky2017WassersteinG}. Then, we briefly introduce our RL model for traffic signal control, based on a state-of-the-art method named PressLight~\cite{wei2019presslight}. Finally, to improve the generalization ability of the RL-based model, we combine the idea of flow clustering and model-agnostic meta-learning (MAML), and propose the enhanced meta reinforcement learning TSC algorithm \our. We will illustrate each of these three modules in the following sections.

\subsection{WGAN-based Traffic Flow Generator}
Due to the difficulty of collecting real data, there are only eleven different traffic flows in one of our datasets, which is obviously insufficient. Since traffic flow in the real world is always changing,  we need to make sure our model can adapt to the traffic flows that are different from the traffic flows they have experienced when training. Therefore, we decide to use WGAN to generate more traffic flows given the collected ones. The fundamental idea of WGAN is to  optimize the following two loss functions alternatively:
\begin{equation}
    \begin{split}
        L_{d} &= \mathbb{E}_{x\sim D_{g}}[F(x)] - \mathbb{E}_{x\sim D_{r}}[F(x)]~, \\
        L_{g} &= \mathbb{E}_{x\sim D_{r}}[F(x)] - \mathbb{E}_{x\sim D_{g}}[F(x)] = - \mathbb{E}_{x\sim D_{g}}[F(x)]~.\\
    \end{split}
    \label{eq:wgan_losses}    
\end{equation} 
where $L_{d}$ and $L_{g}$ denote the loss function of the discriminator and the generator respectively, $D_{r}$ and $D_{g}$ denote the distribution of real data and generated data respectively, $F$ is the neural network to be optimized. Note that in the loss function of the generator, the first term is removed since it is not related to the generator.

In our implementation, the output of the generator will be the generated traffic flows, with the format defined in Equation~\ref{eq:flow}. Besides, we have made some modifications to make the model meet our requirements:

\subsubsection{\textbf{Proper Wasserstein Distance}}
The objective of WGAN is to reduce the Wasserstein distance (W-distance) between generated data distribution and real data distribution. However, in our problem, if the generated traffic flow distribution is too similar to $D_{0}$, there is no way to utilize generated traffic flow distribution to test the generalization ability of models. Therefore, we constrain the W-distance to a fixed number $\epsilon$ by changing the loss function of the generator, as Equation~(\ref{eq:w_dis_loss}) shows. By this new loss function, the generator can generate traffic flow distribution that satisfies its W-distance between $D_{0}$ is a fixed number $\epsilon$.

\begin{equation}
\begin{split}
     L_{g} &= (W(D_{r}, D_{g}) - \epsilon)^{2}= (\mathbb{E}_{x\sim D_{r}}[F(x)] - \mathbb{E}_{x\sim D_{g}}[F(x)] - \epsilon)^{2} \\
    %  & = (\mathbb{E}_{x\sim D_{g}}[F(x)])^{2} - 2\cdot (\mathbb{E}_{x\sim D_{r}}[F(x)] - \epsilon)\cdot \mathbb{E}_{x\sim D_{g}}[F(x)]~.
\end{split}
    \label{eq:w_dis_loss}    
\end{equation}
 
 \subsubsection{\textbf{Other Constraints}}
 Apart from the original Wasserstein distance, we also add more criteria to make the generated traffic flows more real. First, we add a constraint that the amount of vehicles in the generated traffic flow cannot vary more from the one in the dataset, to avoid the generator simply increasing the amount of vehicles to extend the Wasserstein distance. Then, we add another constraint to make sure the variation of vehicle count between two adjacent time intervals cannot be too drastic since vehicle count in the real world cannot vary dramatically in one-time interval. We add these two constraints into the loss function of the generator, represented by $L_{sum}$ and $L_{delta}$. The final loss function of the generator is the combination of all these losses, with $k_{1}$ and $k_{2}$ as hyperparameters of discount factors:
 \begin{equation}
\begin{split}
     L_{g} = (W(D_{r}, D_{g}) - \epsilon)^{2} + k_{1} \cdot L_{sum} + k_{2} \cdot L_{delta}~.
\end{split}
    \label{eq:generator_loss}    
\end{equation}

\subsubsection{\textbf{Implementation of the Discriminator}}
In the original design of WGAN, the neural network of the discriminator $F$ is implemented by several two-dimension convolution layers. However, in our problem, the relation of the vehicle count with the same routes is apparently more important than the relation of vehicle count with different routes. Therefore, we apply one-dimension convolution layers on the vehicle count vector of every route. Then, we let the results of 1-D convolution layers pass fully connected layers to learn the relations between different routes as:
\begin{equation}
\begin{split}
      \text{for}\ m\ in\ &[0\ ,\ |R|) : \\
      \boldsymbol{h}_1^c &= \text{Conv}(\boldsymbol{V}^{m}) \\
      \boldsymbol{e}^{m} &= \text{Conv}(\boldsymbol{h}_1^c) \\
      o =\boldsymbol W_{1} \cdot &~\boldsymbol{e} + \boldsymbol b_{1}~.
\end{split}
    \label{eq:fw}    
\end{equation}
where
the convolution layers are configured with 1x5 filters, the stride of 1, and Sigmoid activation function. 

\begin{figure}[!htbp]
    \centering
    \includegraphics[width=0.45\textwidth]{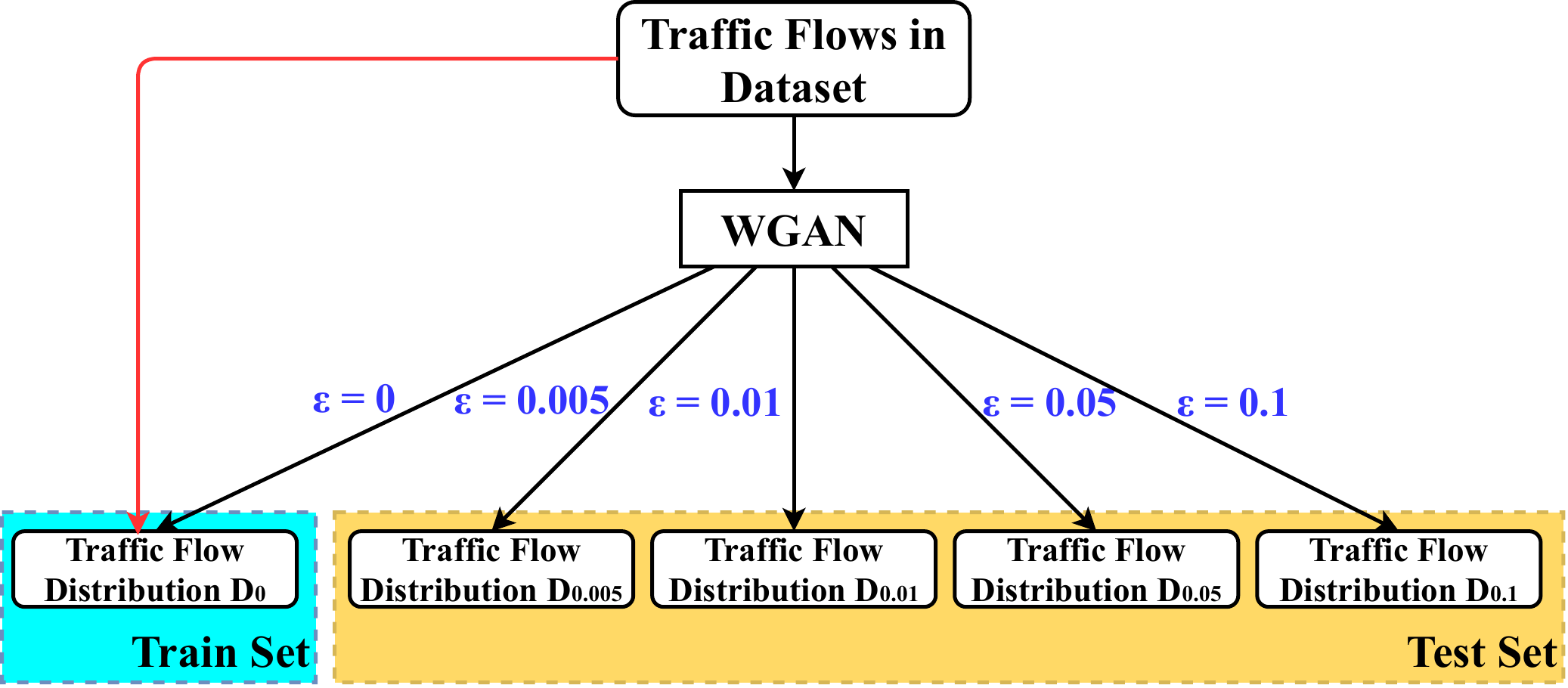}
    \caption{The framework of the WGAN based traffic flow generator. By setting W-distance threshold $\epsilon$ to $0.005, 0.01, 0.05, 0.1$, we can get the traffic flow distributions $D_{0.005}, D_{0.01}, D_{0.05}, D_{0.1}$ to build test traffic environments. Other than the real traffic flows in the dataset, we also utilize the generator to generate more similar traffic flows to build the training environments, by setting $\epsilon$ to $0$.}
    \label{fig:wgan}
\end{figure}

With the help of the modified WGAN, we can generate traffic flow distributions, of which the Wasserstein distance to $D_{0}$ is a fixed number $\epsilon$. By setting $\epsilon$ to $0.005$, $0.01$, $0.05$, $0.1$, we can generate traffic flow distributions $D_{0.005}$, $D_{0.01}$, $D_{0.05}, D_{0.1}$, as the $D_{test}$ required in the problem definition. Note that we also use WGAN to expand the collected real traffic flows using samples from $D_{0}$. Figure~\ref{fig:wgan} shows the framework of our WGAN-based traffic flow generator.

\subsection{RL Model for Traffic Signal Control}\label{sec:rl-tsc}
In this subsection, we introduce the design of the RL agent and learning process with  PressLight~\cite{wei2019presslight} as our backbone algorithm\footnote{PressLight can be regarded as a running RL algorithm in our framework and other RL-based TSC algorithms can be seamlessly incorporated.}. We first introduce the definition of state, action, and reward for the RL agent that controls an intersection. Note that our goal is to improve the generalization ability of TSC models, therefore, our \our framework can be compatible with all TSC models including PressLight.
% \weinan{In fact, for multi-agent setting, we use CoLight instead of PressLight, as claimed in Section~\ref{sec:exp-hangzhou-44}. Shall we make it clear here?}

\begin{itemize}
    \item \textbf{State (Observation).} Our state includes the current phase $\phi$, the number of vehicles on each outgoing lane $x(l_{2}) (l_{2} \in L_{out} )$, and the number of vehicles on every incoming lane $x(l_{1}) (l_{1} \in L_{in})$. Note that the definition of state can be apply to multi-intersection scenario, which equals to the definition of observation in multi-agent RL.
    \item \textbf{Action.} The action of each agent is to choose the phase for the next time interval. In this paper, each agent has eight permissible actions, corresponding to eight phases that are generally accepted and reasonable in previous work. \cite{zheng2019learning}.
    \item \textbf{Reward.} We define the reward of each agent as the negative of the pressure on the intersection it controls, as Equation~(\ref{eq:reward}) shows.
     \begin{equation}
\begin{split}
    w(l_{1}, l_{2}) &= \frac{x(l_{1})}{x_{max}(l_{1})} - \frac{x(l_{2})}{x_{max}(l_{2})}~, \\
    P_{i} &= \Big|\sum_{(l_{1},l_{2})\in i} w(l_{1}, l_{2})\Big|~, \\
    r_{i} &= - P_{i}~.
\end{split}
    \label{eq:reward}    
\end{equation}
where $x(l)$ is the number of vehicles on lane $l$, $x_{max}(l)$ is the maximum permissible vehicle number on lane $l$, $w(l_{1}, l_{2})$ is the pressure of movement $(l_{1}, l_{2})$, and $P_{i}$ is the pressure of the intersection $i$, defined as the sum of the absolute pressures over all traffic movements. Finally, we define the reward of agent $r_{i}$ as the negative value of $P_{i}$. We hope that the vehicles within the traffic system can be evenly distributed by minimizing $P_{i}$, since the pressure $P_{i}$ indicates the degree of disequilibrium between vehicle density on the incoming and outgoing lanes. It has been justified that this reward definition can ultimately minimize the vehicles' average travel time~\cite{wei2019presslight}. 
\end{itemize}

In this paper, we adopt Deep Q-Network (DQN)~\cite{Mnih2015HumanlevelCT} as the function approximator to estimate the Q-value function. This Q-value function network $f_{\theta}$ 
% \weinan{it looks you abuse $f$ here and $f$ in Eq 3.}
takes state as input and predicts the Q-value for each action, to get a better reward. Similar to prior work, we maintain an experience replay memory $\mathcal M$ that stores the experience of agents, defined as transition $(s_{i}, a_{i}, r_{i}, s_{i}')$ (denoting state, action, reward and state of next step respectively). As the traffic simulator runs, we add new transitions to the replay memory $\mathcal M$ and remove old transitions continuously, and the agents will take samples from the replay memory to update their neural networks.

\begin{algorithm}[t]
\DontPrintSemicolon
\caption{Meta-training Process of \our}
\label{alg:train}
\SetKwInOut{Input}{\textbf{Input}}\SetKwInOut{Output}{\textbf{Output}}
\KwIn{
Traffic flow distribution $D_{0}$; step sizes $\alpha$, $\beta$, $\eta$;  number of clusters $C_{n}$; frequency of updating clustering $C_{u}$; total simulate time $T$;}
\KwOut{
Set of optimized parameters initialization $\{ \theta_{0}^{j} | j \in [0, C_{n}) \}$; optimized parameter for the cluster predictor $\theta_{p}$}
\BlankLine
Randomly initialize parameters $\{ \theta_{0}^{j} | j \in [0, C_{n}) \}$ and $\theta_{p}$\\
Randomly initialize flow-cluster mapping $C_{m}$\\
Sample traffic flows $\mathbf V_{i} \sim D_{0}$ \\
Empty feature collector $S, A, R, T^{t}$ \\
\For{round $\longleftarrow$ 0, 1, 2, \dots}
{
    \For{t $\longleftarrow$ 0, 1, 2, \dots, $T$}
    {
        \For{\textbf{all} $\mathbf V_{i}$}
        {
            Find cluster centroid $j \leftarrow C_{m}(\mathbf V_{i})$ \\
            $\theta_{i} \leftarrow \theta_{0}^{j}$ \\
            Generate and store transitions in $\mathcal M_{i}$ using $f_{\theta_{i}}$\\
            Update $\theta_{i} \leftarrow \theta_{i} - \alpha \nabla_{\theta}\mathcal L(f_{\theta}; \mathcal M_{i}) $ \\
            Generate and store transitions in $\mathcal M^{'}_{j}$ using updated $f_{\theta_{i}}$ \\
            Collect state, action and reward to feature collector $S_{i}, A_{i}, R_{i}$\\
            \If{t == T}
            {
                Collect the average total travel time of all vehicles to feature collector $T^{t}_{i}$
            }
        }
        \For{j $\longleftarrow$ 0, 1, 2, \dots, $C_{n}$}
        {
            Update $\theta_{0}^{j} \leftarrow \theta_{0}^{j} - \beta\nabla_{\theta}\mathcal L(f_{\theta}; \mathcal M^{'}_{j})$
        }
    }
    \If{round \% $C_{u} == C_{u} - 1$}
    {
        Use $\langle S, A, R, T^{t} \rangle$ as features to do clustering, tailor all $\mathbf V_{i}$ to $C_{n}$ cluster centroids \\
        Update flow-cluster mapping $C_{m}$ \\
        Update parameter of the cluster predictor  $\theta_{p}\leftarrow \theta_{p}-\eta\nabla_{\theta}\sum_{i} \mathcal L(f_{\theta}(\langle S_{i}, A_{i}, R_{i}\rangle), C_{m}(\mathbf V_{i}))$ \\
        Empty feature collector $S, A, R, T^{t}$ 
    }
}
\end{algorithm}

\subsection{\our Framework}
\begin{figure*}[!htbp]
    \centering
    \includegraphics[width=0.8\textwidth]{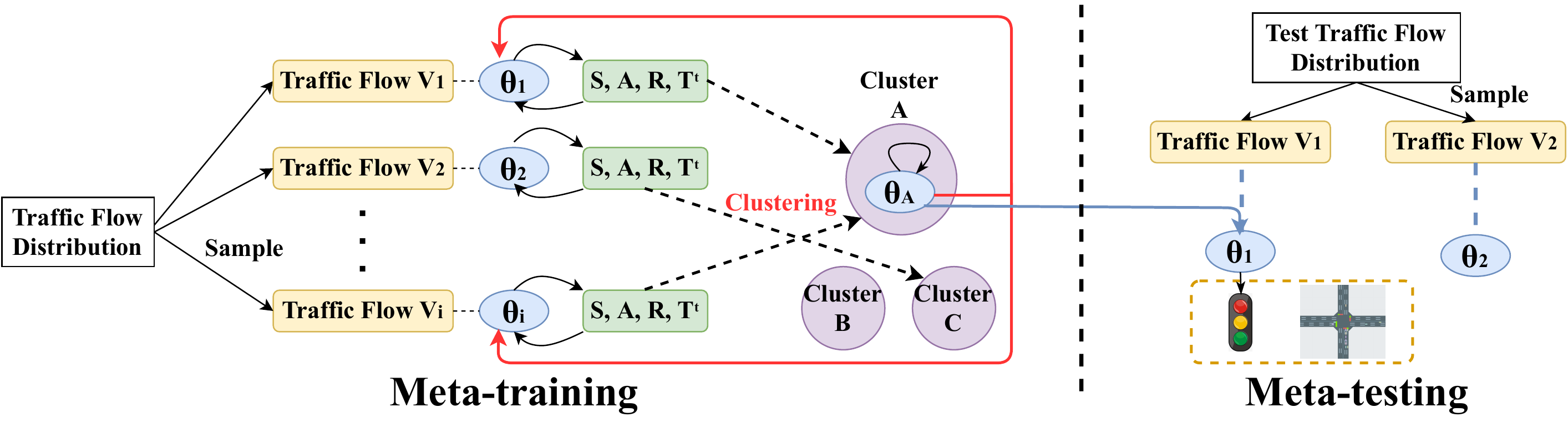}
    \caption{The framework of the proposed \our.}
    % \caption{The framework of the proposed \our. (1) In the meta-training process, we first sample some traffic flows from the given distribution. We feed the every traffic flow to a traffic environment and train Q-network $f_{\theta_{i}}$ in every environment. Then, we use the pattern of <state, action, reward, total travel time> as features to perform clustering. For every traffic flow environment,  we use the experience generated by the updated Q-network $f_{\theta_{i}}$ to update the global parameter initialization  $f_{\theta_{A}}$ in corresponding cluster. Finally, we use $f_{\theta_{A}}$ to replace $f_{\theta_{i}}$ and repeat these steps. (2) In the meta-testing process, we first set up traffic flow environments. For every traffic flow environment, we let the agents randomly explore for a short warm-up period, to get the pattern of <state, action, reward, total travel time> in this new environment. According to its pattern, we assign it with a suitable global parameter initialization, which is the result of the meta-training process.}
    \label{fig:meta-train}
\end{figure*}

Finally, we introduce our \our framework, which reuses previously learned knowledge to adapt to different traffic environments. \our follows the idea of the gradient-based meta-reinforcement learning framework MAML, and its variants Multimodal Model-Agnostic Meta-Learning (MUMOMAML)~\cite{Vuorio2018TowardMM} and Hierarchical Structured Meta-learning
(HSML)~\cite{Yao2019HierarchicallySM}. In the original design, MAML aims to find a proper global initialization $\theta_{0}$ that can adapt to all environments in the distribution. However, some researchers find that only one global initialization is not always enough for all complicated environments. They propose the idea to tailor global initialization to each specific task. This situation also holds in traffic signal control problems, where traffic flow may vary a lot from regular time to rush hours, from one region to another. To overcome this issue, \our utilizes the idea of clustering to train a suitable parameter initialization with specific traffic flow environments, while maintaining a set of global parameter initializations. It turns out the clustering idea does improve the performance compared to the origin MAML in our experiments. The whole process of \our is described in Algorithm~\ref{alg:train} for meta-training, and Algorithm~\ref{alg:test} for meta-testing. The framework of \our is shown in Figure~\ref{fig:meta-train}. We will explain the details in the following subsections.

\subsubsection{\textbf{Flow Clustering.}}
\label{sec:cluster}
When we try to apply MAML to the traffic signal control model, we find that different traffic flows fed to the traffic environment will make different influences on the experience acquired by RL agents, especially the patterns of state, action, and reward. In addition, we notice that the average travel time of vehicles will intuitively reflect the complexity of the traffic flow environment. With the help of these patterns, we think it is worth trying to recognize traffic flows and tailor them to different parameter initializations. 

We design four feature collectors $S, A, R, T^{t}$, to collect state, action, reward, and average travel time features when training with each traffic flow. When clustering, we use k-means~\cite{MacQueen1967SomeMF} which execute three steps alternatively: First, randomly initialize global parameters $\{ \theta_{0}^{j} \}$, and flow-cluster mapping $C_{m}$; Second, after every $C_{u}$ training rounds, calculate the average feature of all feature collectors in every cluster $j$, and use the average feature as the feature of the cluster centroid; Third, for every traffic flow, calculate the distance between its features and the features of all cluster centroids. It will choose the nearest one as its new cluster, and update the flow-cluster mapping $C_{m}$.
\begin{algorithm}[tbp]
\DontPrintSemicolon
\caption{Meta-testing Process of \our}
\label{alg:test}
\SetKwInOut{Input}{\textbf{Input}}\SetKwInOut{Output}{\textbf{Output}}
\KwIn{Traffic flow Distribution $D_{test}$ for testing; step size $\alpha$; Set of learned initialization $\left\{ \theta_{0}^{j} \right\}$ and $\theta_{p}$; total simulate time $T$; learning start time $T_{s}$}
\BlankLine
Sample traffic flows $\mathbf V_{i} \sim D_{test}$ \\
Empty feature collector $S, A, R$ \\
\For{\textbf{all} $\mathbf V_{i}$}
{
    \For{t $\longleftarrow$ 0, 1, 2, \dots, $T_{s} $, \dots, $T$}
    {
        \If{t < $T_{s}$}
        {
            Randomly choose action \\
            Collect state, action and reward to feature collector $S_{i}, A_{i}, R_{i}$\\
        }
        \If{t == $T_{s}$}
        {
            By cluster predictor, tailor $\mathbf V_{i}$ to cluster center $j \leftarrow f_{\theta_{p}}(<S_{i}, A_{i}, R_{i}>)$ \\
            Update parameter initialization $\theta_{i} \leftarrow \theta^{j}_{0}$
        }
        \If{t > $T_{s}$}
        {
            Generate and store transactions in $\mathcal M_{i}$ using $f_{\theta_{i}}$\\
            Update $\theta_{i} \leftarrow \theta_{i} - \alpha \nabla_{\theta}\mathcal L(f_{\theta}; \mathcal M_{i}) $ \\
        }
    }
}
\end{algorithm}
\subsubsection{\textbf{Meta-Training Process}} As we mentioned before, our RL agent has a Q-value network $Q(s,a)$ to be optimized. As the training progresses, the experience of agents will be stored in the replay memory $\mathcal M_{i}$, separated by each traffic flow.

During the meta-training process, with each traffic flow $\mathbf V_{i}$ fed to the traffic environment, the parameter initialization $\theta_{i}$ is set by the global parameter initialization $\theta_{0}^{j}$ of the corresponding cluster $j$. We maintain a flow-cluster mapping $C_{m}$ to find the corresponding cluster given a traffic flow. Then, the parameter $\theta_{i}$ is updated at every several timesteps by gradient descent, taking one gradient step as exemplary: 
\begin{equation}
    \theta_{i} \leftarrow \theta_{i} - \alpha \nabla_{\theta}\mathcal L(f_{\theta}; \mathcal M_{i}) ~,
    \label{eq:update_local}
\end{equation}
where $\alpha$ is the step size, $\mathcal L$ is the loss function to optimize $\theta_{i}$:
\begin{equation}
    \begin{split}
    &\mathcal L(f_{\theta}; \mathcal M_{i}) = \\ &\mathbb E_{s,a,r,s'\sim \mathcal M_{i}}\Big[\Big (r + \gamma \max_{a'} Q(s', a'; f_{\theta_{i}^{-}}) - Q(s, a; f_{\theta_{i}})\Big)^{2}\Big]~,  
    \end{split}
\end{equation}
where $\gamma$ is the discount factor for future reward, and $\theta_{i}^{-}$ are the parameters of target Q-value network. 

After updating $\theta_{i}$, we use the updated neural network $f_{\theta_{i}}$ to generate and store new experience in the replay memory $\mathcal M_{j}^{'}$ of the corresponding cluster. Then, we use the newly sampled transitions to update the global parameter initializations in each cluster, with $\beta$ as step size. Take one gradient step as exemplary: $\theta_{0}^{j} \leftarrow \theta_{0}^{j} - \beta\nabla_{\theta}\mathcal L(f_{\theta}; \mathcal M'_{j})$.
%\begin{equation}
%    \theta_{0}^{j} \leftarrow \theta_{0}^{j} - \beta\nabla_{\theta}\mathcal L(f_{\theta}; \mathcal M'_{j})
%\end{equation}

\subsubsection{\textbf{Meta-Testing Process}} 

In the meta-testing process, we sample new traffic flows from the traffic flow distributions in test set, and feed the new traffic flows to the traffic environment. Then, we put the trained agents in the new traffic environments. However, a problem raised that how to tailor the new traffic flow to existing global parameter initializations in the meta-training process. To solve this problem, we design a cluster predictor, which contains a neural network defined as $f_{p}$. The cluster predictor will predict cluster index based on the features in feature collectors $S, A, R$. We will update the parameter of predictor $\theta_{p}$ during the meta-training process, according to the results of k-means every time. Therefore, we can get a well-trained predictor after the meta-training process via Equation~(\ref{eq:predictor}), with $\eta$ as step size. 
\begin{equation}
    \theta_{p}\leftarrow \theta_{p}-\eta\nabla_{\theta}\sum_{i} \mathcal L\big(f_{\theta}(\langle S_{i}, A_{i}, R_{i}\rangle), C_{m}(\mathbf V_{i})\big)~.
    \label{eq:predictor}
\end{equation}
Note that, we only train the predictor in the meta-training process. We admit that the network of the predictor may be unstable at the beginning of the training process. However, it turns out that the predictor will reach convergence after the clustering mapping become stable in our experiments.

Then, we can utilize this predictor to tailor traffic flows in the test set. In general, when we start to train agents in a new environment, we need to let the agent explore randomly for a short period of time to better adapt to the new environment \cite{Zang2020MetaLightVM}. Therefore, we can take advantage of this warm-up period. During this period, we use feature collectors to collect state, action and reward features in the new environment. When learning start time ends, we feed features in $S, A, R$ to the cluster predictor, get the suitable cluster index $j$, and use $\theta_{0}^{j}$ to initialize the parameters of agents. Then, we use Equation~(\ref{eq:update_local}) again to update the parameters. 
\section{Experiment}
\subsection{Environment settings}
We conduct experiments on Cityflow\footnote{https://github.com/cityflow-project/CityFlow}~\cite{zhang2019cityflow}, an open-source traffic simulation platform. Specifically, Cityflow is used as the environment to provide the state and reward of the traffic control agents and execute the action of agents by changing the phase of traffic lights. 
We will feed the traffic flows sampled from traffic flow distributions to Cityflow in each experiment. Our code has been released on a GitH
ub repository\footnote{https://github.com/only-changer/GeneraLight}.
\subsection{Datasets}
We use three public real-world datasets\footnote{https://traffic-signal-control.github.io/\#open-datasets}\cite{Wei2019ASO, zheng2019learning, wei2019colight}. Roadnets are abstracted from OpenStreetMap\footnote{\url{https://www.openstreetmap.org}}, and traffic flows are extracted from observed real-world data such as surveillance cameras data and vehicle trajectory data. Specifically, the first dataset contains one intersection in Hangzhou (1x1); the second dataset contains five intersections in Atlanta (1x5); the third dataset contains 16 intersections in Hangzhou (4x4).

\subsection{Compared Methods}
To evaluate the generalization ability and effectiveness of our proposed model, we compare our method with the following methods:
\begin{itemize}
    \item \textbf{\dqn} \cite{Mnih2015HumanlevelCT} simply applies the \dqn framework in the traffic signal control problem. It uses the positions of vehicles on the road as state and combines several measures into the reward function.
    \item \textbf{\intelli} \cite{wei2018intellilight} is also a \dqn-based method, which uses richer representations of the traffic information in the state and reward function. In addition, it is optimized for the single intersection scenario by a well-designed neural network.
    \item \textbf{\ddpg}~\cite{Lillicrap2015ContinuousCW} uses a different way to solve the TSC problem, by policy gradient. Instead of optimizing Q-value function, it directly optimizes the policy by an actor-critic based neural network.
    % It has the design of the target network for better training.
    \item \textbf{\frap}~\cite{zheng2019learning} is an efficient reinforcement learning-based method, which is designed to learning the inherent logic of the traffic signal control problem, called phase competition. By merging similar transactions in exploration regardless of the intersection structure and the local traffic situation, \frap has become the best method in the single intersection roadnet.
    \item \textbf{\press}~\cite{wei2019presslight},
    as described in Section~\ref{sec:rl-tsc}, is the state-of-the-art RL based method that can be applied to both single and multi-intersection scenarios. It uses the current phase, the number of vehicles on outgoing lanes, and the number of vehicles on incoming lanes as the state, and uses the pressure as the reward.
    \item \textbf{\colight}~\cite{wei2019colight}
    is designed for the multi-intersection scenario. It uses the attention mechanism to represent the information of neighbors, to achieve the goal of cooperative traffic signal control. This method is the state-of-the-art multi-intersection traffic signal control algorithm.
    \item \textbf{\maml}~\cite{Finn2017ModelAgnosticMF}
    combines each of the above algorithms and the framework of MAML reinforcement learning. In each episode, this method will alternatively update the parameter for each environment and the global parameter initialization.
    \item \textbf{\metalight}~\cite{Zang2020MetaLightVM}
    is designed for transferring knowledge between different intersections, such as training an agent on a four-way intersection and testing it on a five-way intersection. \metalight focuses on the different action spaces and state spaces instead of traffic flow. It designed two-step adaption for solving its problem, called Individual-level Adaptation and Global-level Adaptation.
    \item \textbf{\our}: This is our proposed method in Section~\ref{sec:method}. 
    \item \textbf{\noblock}: A lower-bound of average travel time of a given road network and flow. It is computed assuming all vehicles in the flow travels from origin to destination without being stopped by red lights or blocked by other vehicles. 
\end{itemize}

Note that for the first six methods \dqn, \intelli, \ddpg, \frap, \press, and \colight, they are not designed for training and testing in different traffic environments, which makes all these methods perform not well in our experiment settings. For fairness, we add some modifications to these methods, such as add a shared replay memory for different environments. Besides, we combine MAML to the first six methods to improve their generalization ability. We test these methods with and without the combination of MAML in our experiments.

\begin{figure}[!htbp]
% \vspace{-0.3cm}

% \setlength{\belowcaptionskip}{-0.6cm}
    \centering
    \includegraphics[width=0.45\textwidth]{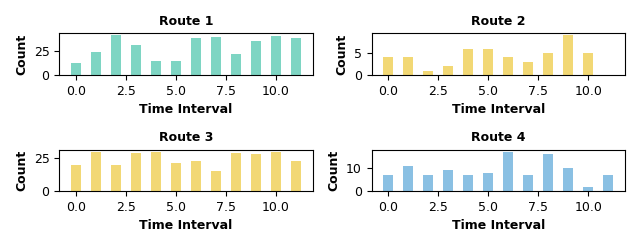}
    \caption{Sample of traffic flow distribution $D_{0}$ in Hangzhou 1x1 dataset, with 10 minutes as time interval.}
    \label{fig:real_flow}
\end{figure}
\begin{figure}[!htbp]
    \centering
    \begin{tabular}{cc}
    \includegraphics[width=0.215\textwidth]{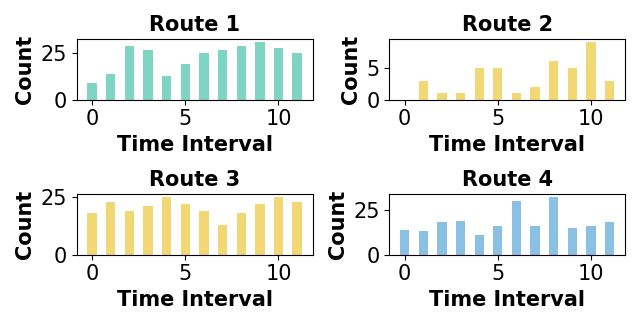} & \includegraphics[width=0.215\textwidth]{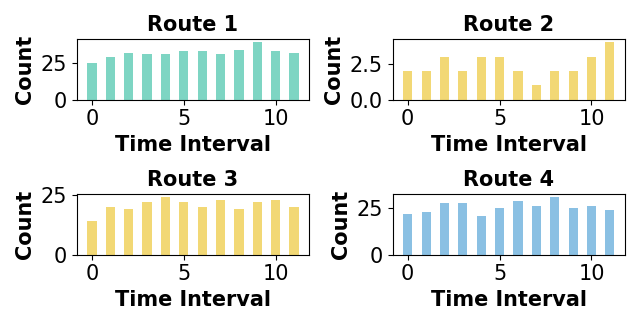} \\
    (a) W-dis = 0.005 &  (b) W-dis = 0.01\\
    \includegraphics[width=0.215\textwidth]{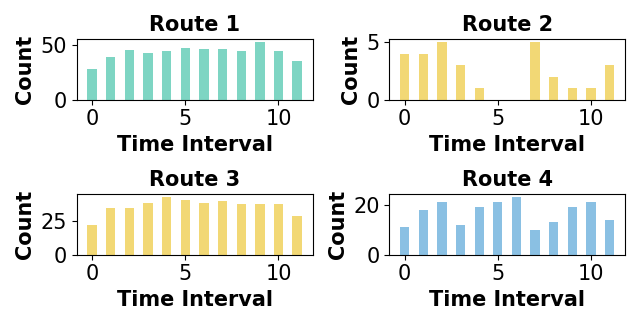} & \includegraphics[width=0.215\textwidth]{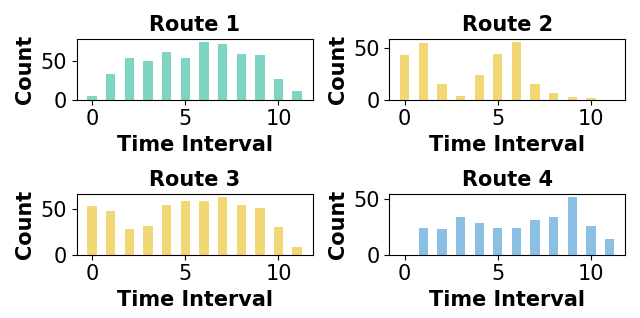} \\
    (c) W-dis = 0.05 &  (d) W-dis = 0.1\\
    \end{tabular}
    \caption{Samples of generated traffic flow distributions $D_{0.005}, D_{0.01}, D_{0.05}, D_{0.1}$, with 10 minutes as time interval.}
    \label{fig:fake_flow}
\end{figure}

\begin{table*}[!thbp]

% \large

\centering
\caption{Overall performance comparison on Hangzhou 1x1 roadnet, where $D_{0}$ is the traffic flow distribution for training, $D_{0.005}, D_{0.01}, D_{0.05}, D_{0.1}$ are the generated traffic flow distributions for testing (subscripts show the Wasserstein Distance between $D_{0}$). Average travel time is reported in the unit of second. ``Improvement'' under \our shows the improvement over the best baseline. ``Improvement'' in the last row shows the relative improvement with respect to \noblock in Equation~(\ref{eq:rel-improve}).}
\label{tab:performance_hz_1x1} 
\resizebox{\textwidth}{!}{
\begin{tabular}{llll|lll|lll|lll|lll}
\toprule & \multicolumn{3}{c}{$D_{0}$} 
       & \multicolumn{3}{c}{$D_{0.005}$, W-dis = 0.005} & \multicolumn{3}{c}{$D_{0.01}$, W-dis = 0.01} &
       \multicolumn{3}{c}{$D_{0.05}$, W-dis = 0.05}   &
       \multicolumn{3}{c}{$D_{0.1}$, W-dis = 0.1}   
         \\ \cline{2-16}
       & Max & Min & Mean & Max & Min & Mean & Max & Min & Mean & Max & Min & Mean & Max & Min & Mean \\ \midrule
       \dqn & 309.6 & 70.6 & 140.8 & 146.0 &	110.7 & 138.5 & 309.4 & 270.1 & 283.3 & 434.9 & 384.6 & 403.5 & 720.3 & 642.9 & 690.4 \\
       \quad +MAML & 194.8 & 89.5 & 124.9 & 138.5 & 129.9 & 137.3 & 242.4 & 210.1 & 225.3 & 375.2 & 331.6 & 358.1 & 683.3 & 562.6 & 641.1 \\
       \cdashline{1-16}[0.8pt/2pt]
       \intelli & 259.3	& 128.1 &	189.9 &	281.4 &	150.2 &	198.3 &	378.9 &	270.9 &	296.5 &	522.7 &	434.6 &	466.7 &	782.9 &	676.9 &	700.1 \\
        \quad +MAML & 198.2 &	109.5 &	165.8 &	248.6 &	169.3 &	178.0 &	326.9 &	254.0 &	286.6 &	502.2 &	376.5 &	413.2 &	726.4 &	573.2 &	656.8\\
        \cdashline{1-16}[0.8pt/2pt]
        \ddpg & 323.2 & 134.2 &	161.1 &	329.6 &	137.8 &	203.3 &	480.8 &	357.5 &	434.7 &	533.4 &	469.3 &	508.3 &	791.4 &	704.1 &	755.0\\
        \quad +MAML & 240.4 & 141.5 &	153.4 &	315.6 &	156.3 &	187.7 &	497.3 &	339.0 &	380.2 &	483.1 &	435.9 &	460.5 &	817.8 &	684.1 &	711.1\\
        \cdashline{1-16}[0.8pt/2pt] 
        \frap & 215.3 &	96.6 &	137.8 &	215.3 &	99.0 &	145.4 &	296.0 &	118.2 &	168.6 &	404.4 &	101.1 &	244.1 &	\textbf{575.3} &	\textbf{485.6} &	\textbf{530.0}\\
        \quad + MAML & 184.3 &	72.6 &	109.6 &	246.7 &	97.9 &	144.7 &	356.8 &	111.7 &	190.4 &	\textbf{327.2} &	\textbf{140.0} &	\textbf{216.7} &	614.3 &	498.1 &	545.5\\
        \cdashline{1-16}[0.8pt/2pt] 
        \press & 277.1 &	70.3 &	127.4 &	180.9 &	126.2 &	135.5 &	307.0 &	277.0 &	293.0 &	361.8 &	314.8 &	331.1 &	743.3 &	634.3 &	705.9\\
        \quad +MAML & 252.8 &	89.5 &	119.5 &	168.3 &	117.0 &	137.2 &	201.0 &	159.5 &	178.8 &	236.7 &	190.4 &	220.6&	589.8 &	514.3 &	560.8\\
        \cdashline{1-16}[0.8pt/2pt]
        \metalight & \textbf{176.8} &\textbf{74.9} &	\textbf{109.3} &	\textbf{134.1} &	\textbf{105.9} &	\textbf{126.9} &	\textbf{195.3} &	\textbf{132.0} &	\textbf{156.6} &	276.5 &	156.0 &	226.1 &	619.1 &	567.5 &	604.5\\
        \midrule
       \our & \textbf{156.4} & \textbf{72.3} & \textbf{106.1} &\textbf{90.9}  & \textbf{86.4}  & \textbf{88.8} & \textbf{112.0} & \textbf{100.0} & \textbf{105.5} & \textbf{140.4} & \textbf{105.6} & \textbf{118.1} & \textbf{492.6} & \textbf{423.8} & \textbf{459.2}\\
       Improvement & 11.6\% &	3.5\% &	2.9\% &	32.2\% &	18.4\% &	30.0\% &	42.7\% &	24.2\% &	32.6\% &	57.1\% &	24.6\% &	45.5\% &	14.4\% &	12.7\% &	13.4\%\\
       \cdashline{1-16}[0.8pt/2pt]
       \noblock & 54.0 &	54.0 &	54.0 &	54.0 &	54.0 &	54.0 &	54.0 &	54.0 &	54.0 &	54.0 &	54.0 &	54.0 &	54.0 &	54.0 &	54.0 \\
       Improvement & 16.6\% &	12.5\% &	5.8\% &	54.0\% &	37.6\% &	52.2\% &	59.0\% &	41.0\% &	49.8\% &	68.4\% &	40.1\% &	60.6\% &	15.9\% &	14.3\% &	14.9\% \\
\bottomrule  
\end{tabular}}
\end{table*}
\subsection{Evaluation Metrics}
\label{sec:metric}
We choose the average travel time as our evaluation metric, which is the most frequently used measure to judge the performance in the traffic signal control problem~\cite{zheng2019learning,wei2019colight,wei2019presslight}. It is calculated by the average travel time of all vehicles spending from origin to destination (in seconds). For every traffic flow distribution, we calculate the max, min, and mean average travel time in all traffic environments with traffic flows sampled from it. Notice that we show two kinds of improvement in our result, one is calculated using raw average travel time, and the other one (relative improvement) is calculated with respect to \noblock as
\begin{equation}
\begin{split}
      \Delta = & \frac{(baseline-\noblock)-(\our-\noblock)}{baseline-\noblock} \\
      = & \frac{baseline-\our}{baseline-\noblock}~. \\
\end{split}
\label{eq:rel-improve}
\end{equation}

Since no TSC method can perform better than the lower bound \noblock, the relative improvement serves as a more accurate measurement for the true performance boost.

\subsection{Generated Flow Distributions}

In the flow generation algorithm, we regard collected real traffic flows as the samples of traffic distribution $D_{0}$. Since we only have eleven real traffic flow in our dataset, we utilize the idea of data augmentation. Specifically, we shuffle some traffic flows by exchanging the flows in some routes. We also manually make some synthetic traffic flows, such as Poisson, Uniform and Gaussian distribution and mix them in the origin traffic flows.
 
Generated flow distributions are important to verify the generalization ability of methods. Therefore, we first take a look at the results of the WGAN-based flow generator. We use generated traffic flows and real traffic flows in the Hangzhou 1x1 dataset as demonstrations. We show samples from $D_{0}, D_{0.005}, D_{0.01}, D_{0.05}, D_{0.1}$ in Figure~\ref{fig:real_flow} and Figure~\ref{fig:fake_flow}. For simplicity, the sampled traffic flow contains four different routes and twelve time intervals. We can see that the gap between the generated traffic flow and the real traffic flow significantly increases as the W-distance increases.

\begin{table*}[!thbp]

% \large

\centering
\caption{Overall performance comparison on Atlanta 1x5 roadnet. Definitions are the same as in Table~\ref{tab:performance_hz_1x1}.}
\label{tab:performance_at_1x5} 
\resizebox{\textwidth}{!}{
\begin{tabular}{llll|lll|lll|lll|lll}
\toprule & \multicolumn{3}{c}{$D_{0}$} 
       & \multicolumn{3}{c}{$D_{0.005}$, W-dis = 0.005} & \multicolumn{3}{c}{$D_{0.01}$, W-dis = 0.01} &
       \multicolumn{3}{c}{$D_{0.05}$, W-dis = 0.05}   &
       \multicolumn{3}{c}{$D_{0.1}$, W-dis = 0.1}   
         \\ \cline{2-16}
       & Max & Min & Mean & Max & Min & Mean & Max & Min & Mean & Max & Min & Mean & Max & Min & Mean \\ 
       \midrule
       \dqn &301.7 &	255.5 &	280.3 &	297.1 &	241.8 &	266.1 &	577.1 &	349.1 &	399.9 &	690.1 &	375.5 &	441.9 &	991.3 &	578.3 &	755.3\\
       \quad +MAML &264.3 &	201.3 &	226.5 &	237.8 &	194.0 &	214.8 &	517.9 &	299.2 &	344.8 &	641.9 &	345.0 &	414.5 &	983.2 &	525.3 &	742.1\\
        \cdashline{1-16}[0.8pt/2pt]
        \press &472.0 &	207.2 &	286.9 &	477.3 &	198.0 &	277.6 &	561.0 &	297.0 &	346.2 &	631.6 &	383.3 &	430.4 & 968.0 &	684.4 &	859.7\\
        \quad +MAML & 223.0 &	189.5 &	210.1 &	230.9 &	195.8 &	218.5 &	\textbf{370.8} &	\textbf{266.2} &	\textbf{306.4} &\textbf{435.7} &	\textbf{361.9} &	\textbf{394.3 }&	\textbf{994.5} &	\textbf{551.8} &	\textbf{669.8}\\
        \cdashline{1-16}[0.8pt/2pt]
        \metalight & \textbf{206.9} &	\textbf{176.8} &	\textbf{196.5} &	\textbf{258.4} &	\textbf{177.0} &	\textbf{209.6} &	463.1 &	279.9 &	338.9 &	583.3 &	386.3 &	429.9 &	991.2 &	616.5 &	819.2\\
        \cdashline{1-16}[0.8pt/2pt] 
       \colight & 580.7	& 195.5 &	433.3 &	593.6 &	187.1 &	435.2 &	785.4 &	256.6 &	579.1 &	903.4 &	421.9 &	797.5 &	911.1 &	526.5 &	849.2\\
        \quad + MAML &632.6 &	140.9 &	427.3 &	600.0 &	190.2 &	290.4 &	869.4 &	255.1 &	558.7 &	919.4 &	580.5 &	794.0 &	992.9 &	527.1 &	847.3\\
        
        \midrule
       \our & \textbf{194.5} & \textbf{165.1} & \textbf{180.3} &\textbf{200.0}  & \textbf{167.7}  & \textbf{181.6} & \textbf{283.4} & \textbf{232.8} & \textbf{253.9} & \textbf{363.3} & \textbf{298.5} & \textbf{329.0} & \textbf{646.6} & \textbf{409.0} & \textbf{554.2}\\
       Improvement & 6.0\% &	6.6\% &	8.2\% &	13.4\% &	5.3\% &	13.4\% &	23.6\% &	12.5\% &	17.1\% &	16.6\% &	17.5\% &	16.6\% &	35.0\% &	25.9\% &	17.3\%\\
       \cdashline{1-16}[0.8pt/2pt] 
       \noblock & 59.4 &	59.3 &	59.4 &	59.4 &	59.4 &	59.4 &	58.6 &	58.5 &	58.5 &	60.3 &	60.2 &	60.2 &	52.3 &	52.3 &	52.3\\
       Improvement & 8.4\% &	10.0\% &	11.8\% &	29.3\% &	7.9\% &	18.7\% &	28.0\% &	16.1\% &	21.2\% &	19.3\% &	21.0\% &	19.6\% &	36.9\% &	28.6\% &	18.7\% \\
\bottomrule  
\end{tabular}}
\end{table*}

\subsection{Overall Performance}
We train and test our method on all three datasets. \our performs the best on all datasets with all test distributions. Below we take a close look at the experiment on each dataset.
\subsubsection{\textbf{Hangzhou 1x1.}}
In the Hangzhou 1x1 dataset, we compare our method with all baselines except \colight, since \colight is designed for multi-intersections, which is similar to another method \dqn when applied to single intersection roadnet. Baselines combined with MAML are shown by the ``+MAML" rows under the origin baselines. It shows that MAML does increase the generalization ability of methods since all baselines combined with MAML perform much better than original baselines.

We can see that \our achieves the lowest average travel time in all traffic flow distributions. \dqn and \intelli perform similarly in test traffic flow distributions since \intelli is the first proposed RL-based traffic signal control method whose optimization is very limited. \ddpg performs badly compared to other baselines, which may because \ddpg is designed for continuous action space instead of discrete action space in our traffic signal control problem. Though the results of \frap, \press, and \metalight are close to the result of our method in traffic flow distribution $D_{0}$, their performance drops when we increase the distance between test traffic flow distribution and train traffic flow distribution. 

With increasing W-distance of test traffic flow distribution, the performances of all the methods tend to downgrade, while \our maintains decent performance. When the W-distance of test traffic flow distribution is less than 0.05, the performance of \our only fluctuates slightly compared to the performance on $D_{0}$. This has verified the generalization ability of our method facing new environments with different traffic flows.
\begin{table*}[!thbp]

% \large

\centering
\caption{Overall performance comparison on Hangzhou 4x4 roadnet. Definitions are the same as in Table~\ref{tab:performance_hz_1x1}.}
\label{tab:performance_hz_4x4} 
\resizebox{\textwidth}{!}{
\begin{tabular}{llll|lll|lll|lll|lll}
\toprule & \multicolumn{3}{c}{$D_{0}$} 
       & \multicolumn{3}{c}{$D_{0.005}$, W-dis = 0.005} & \multicolumn{3}{c}{$D_{0.01}$, W-dis = 0.01} &
       \multicolumn{3}{c}{$D_{0.05}$, W-dis = 0.05}   &
       \multicolumn{3}{c}{$D_{0.1}$, W-dis = 0.1}   
         \\ \cline{2-16}
       & Max & Min & Mean & Max & Min & Mean & Max & Min & Mean & Max & Min & Mean & Max & Min & Mean \\ 
       \midrule
       \dqn &979.1 &	846.6 &	921.0 &	1016 &	935.7 &	986.7 &	1069 &	964.1 &	1010 &	1127 &	1024 &	1088 &	1284 &	1209 &	1249\\
       \quad +MAML & 530.1 &	394.8 &	448.2 &	605.0 &	462.0 &	529.5 &	707.9 &	504.2 &	617.0 &	855.8 &	715.7 &	781.0 &	1147 &	884.4 &	1021\\
        \cdashline{1-16}[0.8pt/2pt]
        \press &395.9 &	384.6 &	389.9 &	\textbf{435.8} &	\textbf{415.7} &	\textbf{426.6} &	\textbf{484.2} &	\textbf{436.2} &	\textbf{460.3} &	662.0 &	565.7 &	592.9 &	855.1 &	749.2 &	813.4\\
        \quad +MAML & 468.1 &	395.6 &	424.0 &	597.9 &	468.9 &	520.5 &	712.4 &	569.1 &	634.5 &	909.8 &	665.9 &	801.4 &	1100 &	817.0 &	1004\\
        \cdashline{1-16}[0.8pt/2pt]
        \metalight & 514.4 &	475.0 &	497.2 &	587.3 &	512.0 & 	545.6 &	671.6 &	580.5 &	637.8 &	843.3 &	719.9 &	767.4 &	937.9 &	814.2 &	861.1\\
        \cdashline{1-16}[0.8pt/2pt]
       \colight & \textbf{425.6} &	\textbf{368.7} &	\textbf{383.9} &	603.0 &	387.7 &	456.4 &	721.5 &	402.7 &	537.8 &	887.9 &	517.1 &	624.1 &	1040.4 &	726.2 &	880.2\\
        \quad +MAML & 402.7 &	390.9 &	397.5 &	490.4 &	424.8 &	452.0 &	531.7 &	458.6 &	482.9 &	\textbf{643.1} &	\textbf{513.0} &	\textbf{570.2} &	\textbf{785.4} &	\textbf{638.0} &	\textbf{718.3}\\
        \midrule
       \our & \textbf{385.1} & \textbf{376.3} & \textbf{380.0} &\textbf{410.2}  & \textbf{393.6}  & \textbf{402.0} & \textbf{458.7} & \textbf{409.4} & \textbf{432.2} & \textbf{582.0} & \textbf{443.5} & \textbf{493.5} & \textbf{652.2} & \textbf{579.8} & \textbf{622.5}\\
       Improvement	 &	9.5\% &	$\quad \backslash$ &	1.0\% &	5.9\% &	5.3\% &	5.8\% &	5.3\% &	6.1\% &	6.1\% &	9.5\%	 & 13.5\% &	13.5\% &	17.0\% &	9.1\% &	13.3\% \\
       \cdashline{1-16}[0.8pt/2pt]
       \noblock & 332.3 &	331.4 &	331.8 &	332.5 &	331.9 &	332.2 &	330.2 &	329.2 &	329.9 &	339.1 &	338.2 &	338.6 &	350.1 &	349.7 &	349.9\\
       Improvement & 43.5\% &	$\quad \backslash$ &	7.5\% &	24.8\% &	26.3\% &	26.0\% &	16.6\% &	25.0\% &	21.6\% &	20.1\% &	39.8\% &	33.1\% &	30.6\% &	20.2\% &26.0\%\\
\bottomrule  
\end{tabular}}
\end{table*}
\subsubsection{\textbf{Atlanta 1x5.}}
For multi-intersection traffic signal control, we compare \our with \dqn, \press, \metalight, and \colight, with and without MAML in both Atlanta 1x5 dataset and Hangzhou 4x4 dataset. When applying \dqn, \press, and \metalight to the multi-intersection scenario, each agent will be trained individually and regard other agents as part of the environment.

We can see that \our still performs the best in all traffic flow distributions. Note that when finding the best baseline in one traffic flow distribution, we select the one with the lowest mean travel time. Similar to the results in the Hangzhou 1x1 dataset, the improvement of our method is more significant when the W-distance of test traffic flow distribution is larger. When testing on the most challenging traffic flow distribution $D_{0.1}$, \our even achieves a more than 30\% improvement.

\subsubsection{\textbf{Hangzhou 4x4.}}\label{sec:exp-hangzhou-44}
In the Hangzhou 4x4 dataset, we find that the performance of \press is unsatisfactory when applied to 16-agent reinforcement learning. Therefore, we use \colight as the base RL model in \our.

We can see that \our still performs the best. 
% In particular, \our performs very well in the max travel time metric compared to baselines, which means \our outperforms other methods under the hardest environments. 
The performance of \colight is much better than \press since \colight has more optimization for multi-intersections. We notice that the raw improvement in this dataset is not as significant as before. Since the problem of traffic signal control in such a large roadnet is much harder than in 1x1 and 1x5 roadnet, the average travel times of all methods are very long. Though \our has saved more than one hundred seconds compared to baselines, the raw improvement of our method is as high as it supposed to be. However, the relative improvement which indicates real performance boost still remains high, showing the effectiveness of our method.

\begin{figure}[t]
\setlength{\belowcaptionskip}{-0.1cm}
    \centering
    \includegraphics[width=0.4\textwidth]{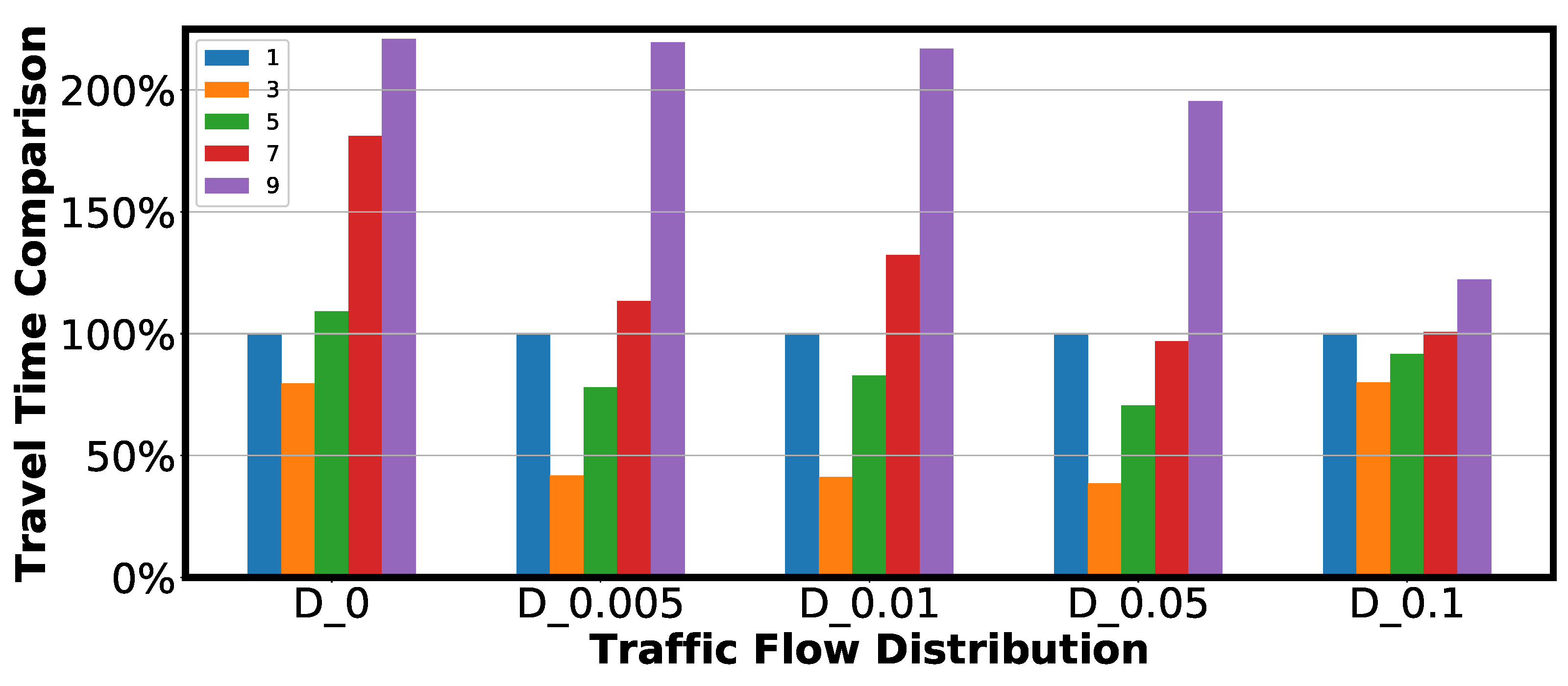} 
    \caption{Performance of \our with respect to different cluster amount. For each test flow distribution, we treat no cluster (cluster number is one) as baseline and show the relative travel time of different cluster by proportion.}
    \label{fig:ab}
    %\vspace{-10pt}
\end{figure}

\subsection{Hyperparameter Study}
The most important hyperparameter in \our is the number of clusters. In Figure~\ref{fig:ab}, we show how cluster amount impacts the performance of \our. We conduct experiments on Hangzhou 1x1 dataset, and use the mean average travel time of all sampled traffic flows as the metric. When setting cluster number to 1, it means we are not performing clustering when training. For each test flow distribution, we treat no cluster (cluster number is one) as our baseline and show the relative travel time of different clusters by proportion.

We can see that \our achieves the best performance with the cluster number as 3, which verifies the necessity of flow clustering. However, as the cluster amount grows from 5 to 9, the performance of \our begins to drop since too many clusters will result in insufficient training samples in each cluster, which leads to underfitting. In general, 3 to 5 clusters are suitable for traffic signal control problems.

\subsection{Case Study}
\begin{figure}[!htbp]
\setlength{\belowcaptionskip}{-0.1cm}
    \centering
    \begin{tabular}{cc}
    \includegraphics[width=0.215\textwidth]{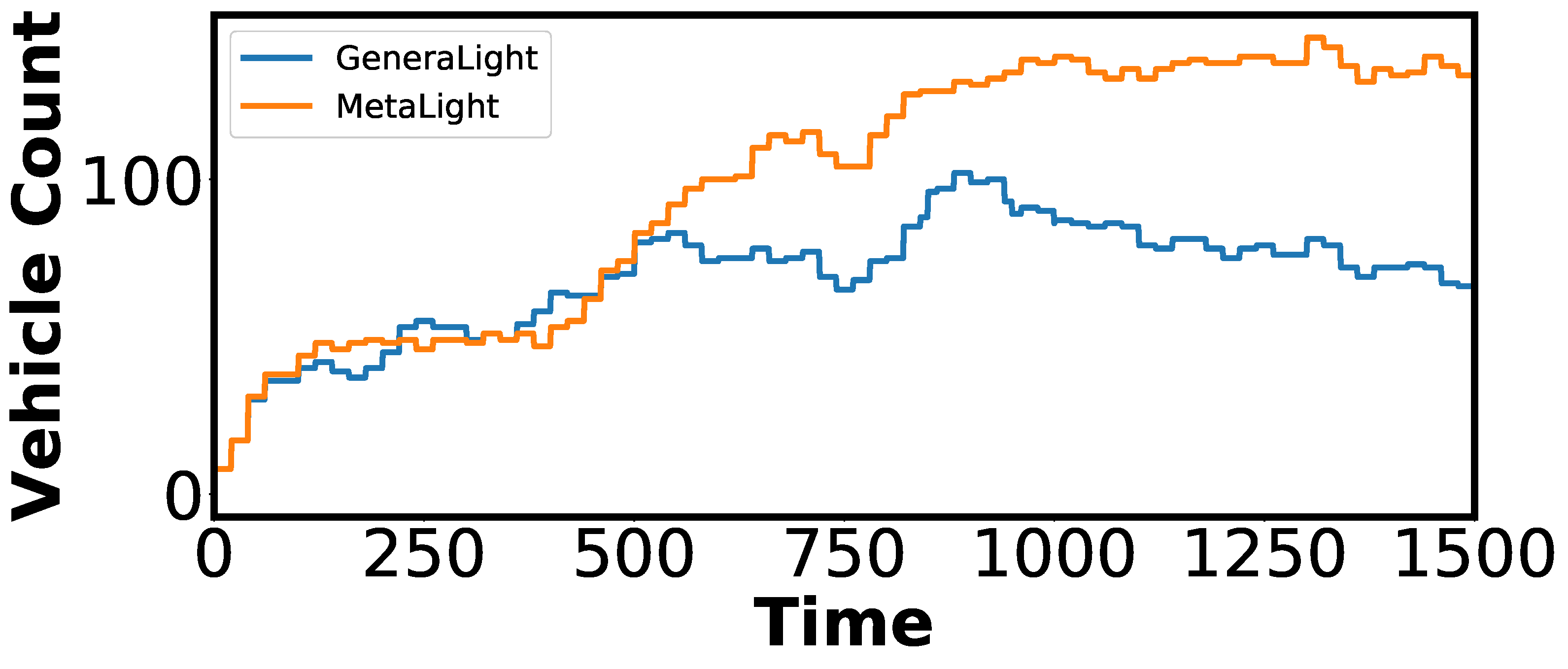} &
    \includegraphics[width=0.215\textwidth]{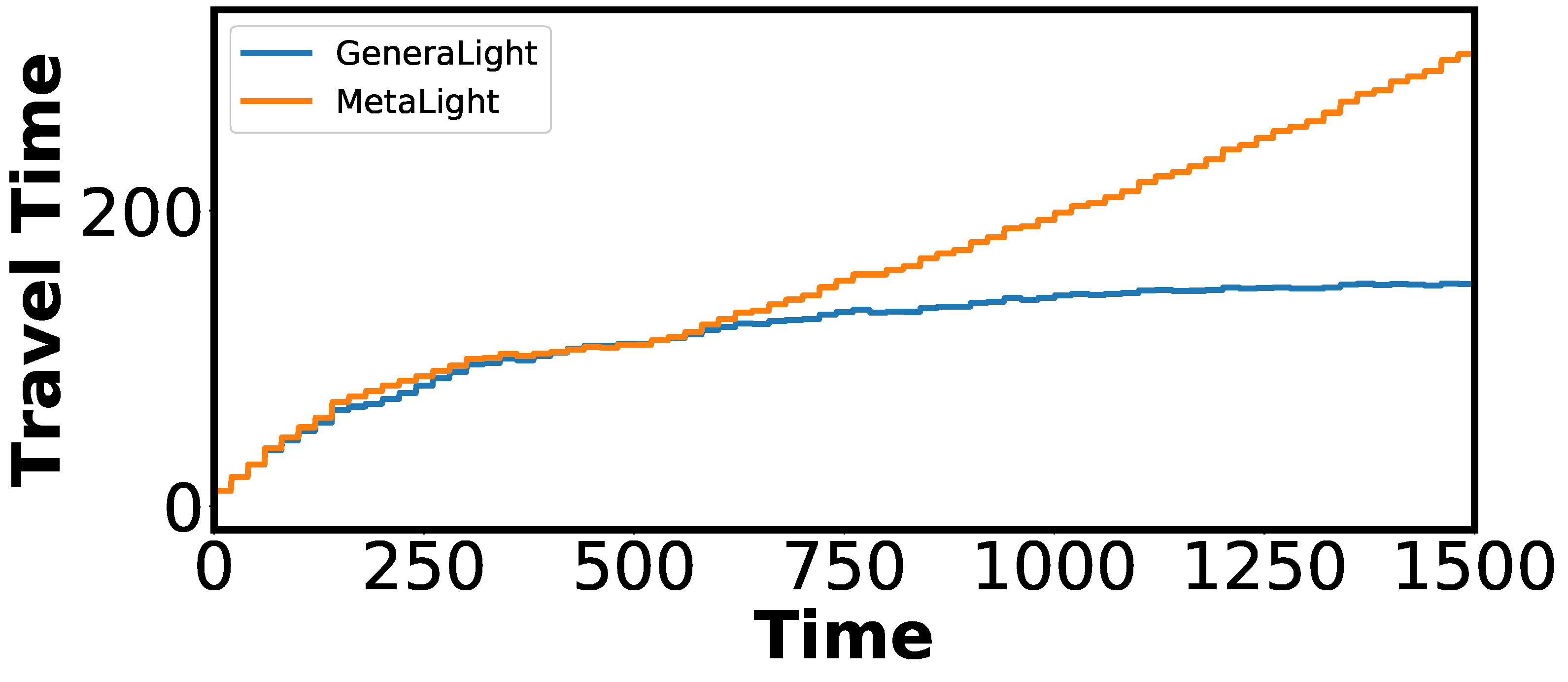} \\ 
  (a) Real-Time Vehicle Count & (b) Real-Time Travel Time \\
  % \includegraphics[width=0.15\textwidth]{metalight_cut.png} & %\includegraphics[width=0.1481\textwidth]{cmaml_cut.png} \\
  % (c) \metalight, 900s & (d) \our, 900s \\
    \end{tabular}
    \caption{Results of Hangzhou 1x1 roadnet, with traffic flow sampled from distribution $D_{0.01}$. We compare \our with the best baseline \metalight. (a) shows the real-time total vehicle count in the whole roadnet. (b) shows the real-time average travel time of all vehicles.}
    \label{fig:case}
\end{figure}

To show the generalization ability of our model, we take a closer look at the traffic control result in Hangzhou 1x1 dataset. As Figure~\ref{fig:case} shows, we choose the traffic flow environment sampled from traffic flow distribution $D_{0.01}$, and compare \our with \metalight, the best baseline in these settings. 
\begin{figure}[!tbp]
\setlength{\belowcaptionskip}{-0.1cm}
    \centering
    \includegraphics[width=0.3\textwidth]{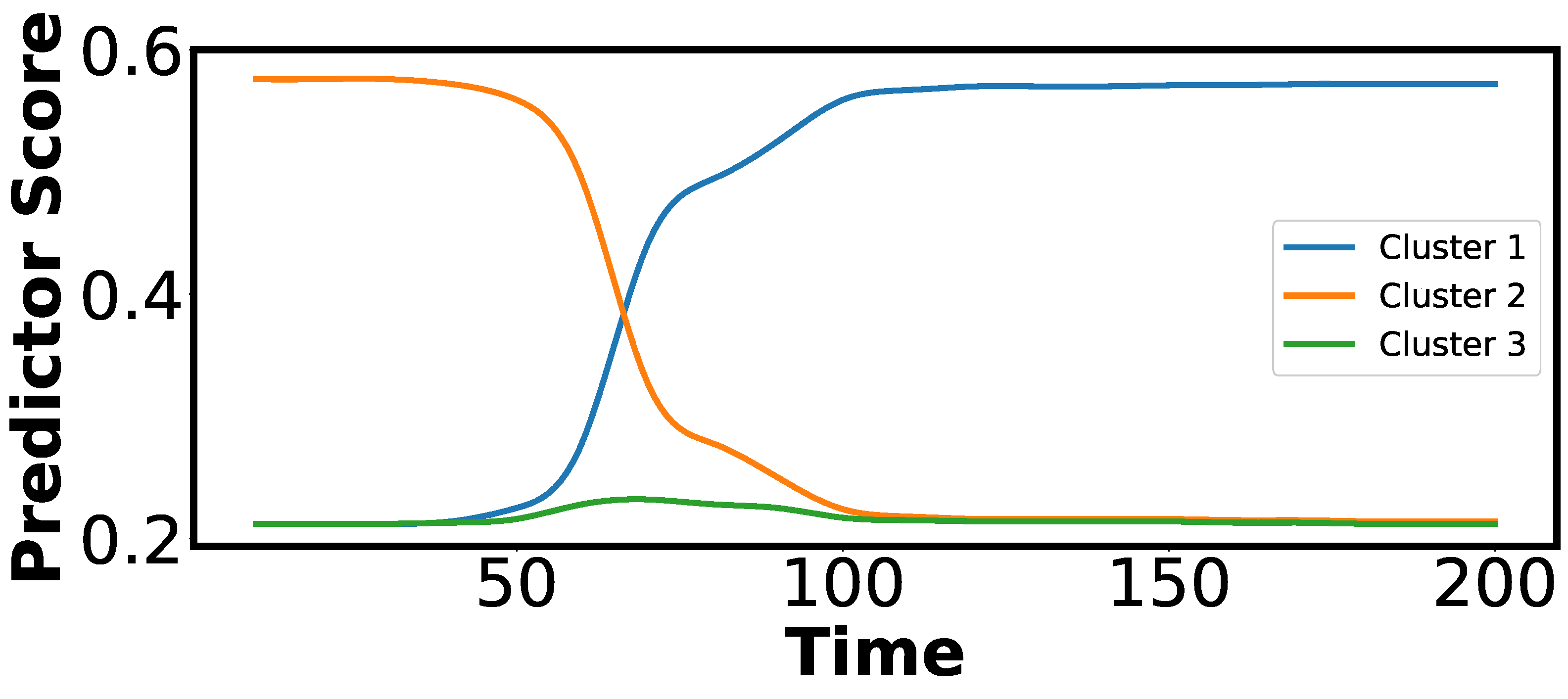} 
    \caption{Probability of each cluster predicted by the cluster predictor during the first 200 steps.}
    \label{fig:predictor}
\end{figure}
We can see that the performances of \our and \metalight are similar at the beginning since the traffic flow is very gradual and steady. However, we can see there are two traffic surges at around 500s and 750s, as Figure~\ref{fig:case}(a) shows. These two traffic surges can be regarded as a form of traffic flow environmental change, which we hope the traffic control model can handle by its generalization ability. In the first traffic surge around 500s, the real-time vehicle count of \our and \metalight both begin to rapidly rise, indicating the traffic congestion occurs. As a result, \our successfully alleviates the congestion as the real-time vehicle count begins to decrease at around 600s, while \metalight fails to handle the congestion and its real-time vehicle count continues to increase. Similar situation happens in the second traffic surge at around 750s. We can see that after two traffic surges, the real-time vehicle count of \our tend to decrease to the previous level, indicating that \our has relieved the traffic congestion twice. In contrast, \metalight fails to deal with traffic congestion and the real-time vehicle count rises to a higher level and never goes down. The same conclusion is also shown in Figure~\ref{fig:case}(b), where the gap between the real-time average travel time of the two methods is clearly widened after the traffic congestion. Above all, \our has proven its generalization ability in this case study. 

We also plot the probability of each cluster predicted by the cluster predictor during the first several hundred steps in Figure \ref{fig:predictor}. We can see that the cluster predictor quickly choose the right cluster using the features collected during the beginning period. Due to the page limit, we do not show the accuracy of the predictor in each experiment. On average, our predictor can reach more than 75\% accuracy. In practice, we can reset the cluster at predefined time interval (e.g. every 4 hours), allowing adjustment for unexpected changes in traffic flows.

\section{Conclusion}
In this paper, we aim at improving the generalization ability of RL-based traffic signal control methods in different traffic flow environment. We first design a WGAN-based traffic flow generator to generate more flows and use them to test the generalization of TSC models. Then, We propose \our, which combines the idea of flow clustering and MAML to improve the generalization ability of RL-based TSC models. \our uses the experience of agents as features to perform clustering and maintains a set of global parameter initializations. Extensive experiments show the superior generalization ability to different traffic flow environments of \our compared to various baselines. For future work, we will try to find more aspects of difference in traffic environments, such as different roadnets, instead of just different traffic flows. 
% We also plan to combine the idea of transfer learning to improve the generalization ability among different cities. 
% In our experiments, we find that training in many different environments simultaneously is time consuming. In the future, we will try to find a way to reduce the training time while maintaining the generalization ability.
% \newpage
\bibliographystyle{ACM-Reference-Format}
\bibliography{reference}
\end{document}